# Feature visualization of Raman spectrum analysis with deep convolutional neural network

*Masashi Fukuhara*[*a], *Kazuhiko Fujiwara*[a], *Yoshihiro Maruyama, Hiroyasu Itoh*

[a] *Tsukuba Research Laboratory, Central Research Laboratory, Hamamatsu Photonics K.K., Ibaraki, Japan*

ABSTRACT

We demonstrate a recognition and feature visualization method that uses a deep convolutional neural network for Raman spectrum analysis. The visualization is achieved by calculating important regions in the spectra from weights in pooling and fully-connected layers. The method is first examined for simple Lorentzian spectra, then applied to the spectra of pharmaceutical compounds and numerically mixed amino acids. We investigate the effects of the size and number of convolution filters on the extracted regions for Raman-peak signals using the Lorentzian spectra. It is confirmed that the Raman peak contributes to the recognition by visualizing the extracted features. A near-zero weight value is obtained at the background level region, which appears to be used for baseline correction. Common component extraction is confirmed by an evaluation of numerically mixed amino acid spectra. High weight values at the common peaks and negative values at the distinctive peaks appear, even though the model is given one-hot vectors as the training labels (without a mix ratio). This proposed method is potentially suitable for applications such as the validation of trained models, ensuring the reliability of common component extraction from compound samples for spectral analysis.

## 1. Introduction

Raman spectroscopy is a non-destructive measurement that is used in various fields such as pharmacy [1], biology [2], and material science [3]. A Raman spectrum has characteristic peaks attributable to the vibrational frequencies of functional groups, which makes it possible to discriminate species of molecules. In general, because of the weakness of the Raman signal, in addition to the preparation of sensitive instruments, various signal processing procedures have been applied [4] to extract usable information from the measured spectra. For instance, various numerical procedures have been proposed: baseline correction [5], noise reduction [6], and statistical procedures such as principal component analysis (PCA) and neural networks [7-9]. PCA and neural networks have been employed as feature extraction techniques to detect components of interest and recognize the target sample. Meanwhile, deep neural networks (DNNs) have also attracted much attention as an extraction technique [10], because significant improvements in classification accuracy have been reported for them in the field of image recognition [11].

DNNs have already been used for object detection [11, 12], image super-resolution [13], and speech recognition systems [14]. Specifically, convolutional neural networks (CNNs) are a well-known model for feature extraction, because they enable the extraction of the feature vector from input data, as with PCA. A CNN model has also been proposed for spectrochemistry [15, 16]. In addition, it was reported that the CNN model has an advantage as an end-to-end learning tool [17]. Although the model has achieved superior performance, it is still difficult to intuitively understand the representations of a deep CNN model.

Explainable artificial intelligence has been proposed and developed to interpret the representations of CNN models [18]. In the field of image classification, a visual explanation method for CNN model has already been reported [19, 20]. The visualization of the features extracted by a CNN is expected to accelerate the development, adjustments, and optimization of the suggested model.

* *Corresponding author.* Tel: +81-29-847-5161; fax: +81-29-847-5266.
  E-mail address: masashi.fukuhara@crl.hpk.co.jp

In this paper, we present a feature visualization for CNN models for Raman spectrum analysis and a common component extraction method from mixed spectra that uses a CNN.

For Raman spectrum analysis, the peak positions (wavenumber or frequency) contain important information that discriminates molecules. Therefore, we first discuss the hyperparameters of a convolutional layer for a CNN designed to extract peak regions using simple spectra generated by a Lorentzian function and white noise. Next, we apply our proposed method to the discrimination of pharmaceutical samples and confirm the Raman peak extraction. Finally, we demonstrate the possibility of common component extraction by evaluating mixed spectra obtained from amino acids.

## 2. Method

### 2.1. CNN recognition model

In this work, the CNN model was designed on the TensorFlow platform [21]. Figure 1 shows the model for the Raman spectrum analysis. We employed a typical hidden layer combination, i.e. a convolutional layer with an activation function, pooling layer, and fully-connected (FC) layer1 [4, 17].

The convolutional layer is described as follows:

$$Y_{x,k'} = f\left(b_{x,k'} + \sum_{k} Cw_{k,k'} * X_{x,k}\right), \quad (1)$$

where $X$ is an input spectrum, $Y$ is the output spectrum, $x$ is the wavenumber index of the input spectrum, $k$ is the index of the input spectrum map, and $k'$ is the index of the output spectrum map. Further, $C$w denotes the convolutional filter weights and $b$ is a bias parameter. Function f indicates an activation function, and * denotes convolution. Stride and padding parameters were chosen so that the output size is identical to the input size. For the subsequent Raman spectrum analysis, the convolutional layers had 64 filters of size 8. These parameters were determined based on the filter dependence of the extracted spectrum regions described in Section 4.1 to extract Raman peaks as if we focused, because it is well known that the peak width and position are important factor for the Raman spectrum evaluation. Leaky ReLUs [22] were used as the activation function at the output of every convolutional layer. This function is described as follows:

$$f(x) = \begin{cases} x & \text{if } x \geq 0 \\ \alpha x & \text{if } x < 0 \end{cases}. \quad (2)$$

In this work, $\alpha$ is 0.2, which is TensorFlow's initial parameter value.

The convolutional layer is followed by a max-pooling layer. The pooling layer performs down-sampling by taking the largest value from its input using a pooling filter. The max-pooling layer is described as follows:

$$Y_{x',k} = \max_{0 \leq m < s}(X_{x \cdot s + m,k}), \quad (3)$$

where $s$ is the pooling filter size and $x$ and $x'$ are wavenumber indices of the input and output spectra, respectively. In this work, 1/2 down-sampling was performed at each pooling layer.

The FC layer was connected between the last pooling layer and the output of the model. The FC layer is described as follows:

$$Y_{k'} = b_{k'} + \sum_{x \cdot k} Fw_{k',x \cdot k}\, X_{x \cdot k}\,, \qquad (4)$$

where $Fw$ denotes the weights of the FC layer. Basically, the input data $X$ were arranged in one dimension because of the flattening layer placed just before the first FC layer. In Fig. 1, the input data $X$ of the layer are the one-dimensionally arranged output of pooling layer 2. A dropout layer was inserted after the first FC layer to avoid the overfitting problem [23]. After the extraction of 128 feature vectors, a second FC layer was connected to the output of the CNN model to derive the predicted value. The softmax function was used just before the output to fit the value within the range 0–1.0.

While training the network, we used cross entropy loss, defined as follows:

$$loss = -\sum_{n} \{y_n^t \log(y_n^p)\}\,, \qquad (5)$$

where $n$ is the label index, and $y^t$ and $y^p$ are the true and predicted values, respectively. The true labels are given by one-hot vectors [24]. The dimensions of $y^t$ and $y^p$ were 9 in the experiments in Section 4.1 and 20 in the experiments in Section 4.2.

To input the measured spectra dataset to the model, its intensity and wavenumber were normalized relative to each maximum value and adjusted in the range of 350–1,800 cm$^{-1}$ with a resolution of 1 cm$^{-1}$ using two-dimensional spline interpolation. Before the interpolation, a linear baseline subtraction was also performed on the measured raw spectra shown as Fig. S1 (supporting information).

### 2.2. Visualization method

Our proposed method derives the "importance weights" mentioned in the approach using the gradients of a target label [20]. The previous work highlights the feature contributing to recognition in images with deep CNN models (AlexNet11 and VGG16 [25]). Using this method for spectrum analysis, the importance is derived by the following equation:

$$\alpha_k^c = \sum_x \frac{\partial y^c}{\partial X_{x,k}}\,, \qquad (6)$$

where $\alpha$ is the importance weight of activation map $X$ (a spectrum), $c$ is the target class, $k$ is the index of activation map $X$, $y^c$ is the predicted score for class $c$, and $x$ is wavenumber index. In this work, we derive the importance weight α by taking the output of the pooling layer for X. Finally, the contributing region L for target class c is calculated by the following equation:

$$L_x^c = ReLU\left(\sum_k \alpha_k^c X_{x,k}\right). \qquad (7)$$

In a deep CNN model, the derived map is well highlighted in the contributing region for image classification using an activation map, however, in shallow layers, only ordinary features, such as edges, were extracted [20]. For Raman spectrum analysis, a shallower model should be able to recognize the spectrum because of the low-dimensional dataset (spectrum) [16, 17]. Here, we propose a method in which the importance spectrum region is derived from the weights in the FC layer. The importance weights of feature vector β are calculated by the following equation:

$$\beta_l^c = \sum_x \sum_k \frac{\partial y^c}{\partial Fw_{x,k,l}} \tag{8}$$

where $c$ is a target class, $x$ is the wavenumber index, $k$ is the index of the feature map (which is output of convolutional or pooling layer), $l$ is the index of the feature vector in the FC layer, $Fw$ denotes the weights of the FC layer (which is just before the output layer), and $y^c$ is the score for class $c$. The importance weights $\beta$ for class $c$ are calculated from the gradient of the score with respect to the weights of the FC1 layer. Then, contribution feature map $M$ is calculated as follows:

$$M_x^c = \sum_k A_{x,k} \left( \sum_l \beta_l^c Fw_{x,k,l} \right), \tag{9}$$

where $A$ is a feature map of the pooling layer. In the case of Fig. 1, 128 importance weights $\beta$ are calculated from each weight of the 128 feature vectors in FC1 using Eq. (8). Then, the importance region focused on by the FC layer [which is indicated in parentheses in Eq. (9)] is calculated from the importance of the feature vector and its weights. Finally, the contribution feature map $M$ is derived by multiplying the calculated important region by the activation map $A$ of the former layer. In this paper, we used linear interpolation to link the down-sampled contribution feature (363 ch) to input size (1451 ch).

In Section 4.1, simple test spectra are evaluated to discuss the filter parameter dependence of the extracted region. The spectra were calculated from a Lorentzian function and white noise as follows:

$$f(x) = \frac{1}{1 + \left\{ \frac{x - x_0}{FWHM/2} \right\}^2} + \sigma(x), \tag{10}$$

where $x$ is the channel (or wavenumber) index, $x_0$ is the peak position, FWHM is the full-width at half maximum of the peak, and $\sigma$ is white noise function. In this work, the value of the white noise was set to be within ± 2.5% of the maximum value of each spectrum.

## 3. Materials and Experiments

### 3.1. Pharmaceutical compound

The following nine pharmaceutical samples were purchased and used in the experiments: Benza® Block® L (Takeda Pharmaceutical Company, Tokyo, Japan), Bufferin A (Lion Corporation, Tokyo, Japan), Eve® A (SSP Co., Ltd., Tokyo, Japan), Rhinitis medication A Kunihiro (Kokando Pharmaceutical Co., Ltd., Hyogo, Japan), Lulu attack® EX (Daiichi Sankyo Co., Ltd., Tokyo, Japan), Norshin AI (ARAX Co., Ltd., Aichi, Japan), Shin jikina® troche (FUJIYAKUHIN Co., Ltd., Saitama, Japan), Pabron rhinitis chewable tablets (Taisho Pharmaceutical Co., Ltd., Tokyo, Japan), and Pabron S α granules (Taisho Pharmaceutical Co., Ltd. Tokyo, Japan). The tablets of the purchased samples were triturated in a mortar before Raman measurement. In this work, the above samples were labelled as BbL, Buf, EvA, Knh, LaE, Njt, Nrs, Pbb, and Psa, respectively.

### 3.2. Amino acid

We prepared 20 amino acids: alanine (Ala), arginine (Arg), asparagine (Asn), aspartic acid (Asp), cysteine (Cys), glutamine (Gln), glutamic acid (Glu), glycine (Gly), histidine (His), isoleucine (Ile), leucine (Leu), lysine (Lys), methionine (Met), phenylalanine (Phe), proline (Pro), serine (Ser), threonine (Thr), tryptophan (Trp), tyrosine (Tyr), and valine (Val). All samples were purchased from Wako Pure Chemical Industries, Ltd. (Osaka, Japan) and triturated in a mortar.

*3.3. Experimental setup*

All Raman spectra were obtained with a Raman spectroscopic module C12710 (Hamamatsu Photonics K.K., Shizuoka, Japan). The prepared samples were measured by focusing near-infrared laser light (wavelength: 785 nm, power: 50 mW). The acquisition time was 1 s per spectrum. The holders for the triturated powder samples were prepared by fixing a silicone rubber with a hole 5 mm in diameter and 1 mm in depth onto a common glass slide S1112 (Matsunami Glass Ind., Ltd., Japan). The samples were placed into the hole of the sample holder.

## 4. Results and Discussion

*4.1. Feature extraction simulation*

First, we prepared simple training and test spectra using Eq. (10) to demonstrate the feature extraction for spectrum recognition. Here, we set three true labels for the prepared Lorentzian spectrum, which are related to the defined peak positions (100, 500, and 1,000 ch). Twenty training spectra were generated for each training label (i.e. a total of 60 spectra) using Eq. (10). Note that the input x-axis size of training and test spectra used in this section was adjusted to 1,024 ch for the evaluation. The network was trained using stochastic gradient descent on an Adam optimizer with a learning rate of $1 \times 10^{-4}$ for 100 epochs.

Figure 2 shows the filter parameter dependences of the extracted feature. The black solid lines indicate the evaluated test spectra. Figure 2(a) shows the filter size dependence. It was confirmed that the extracted region was more focused on the peak position as the filter size decreased from 128 ch to 8 ch. Here, the FWHM was given as 4 ch. Hence, we confirmed that the filter size should be adjusted to be close to the peak width to extract a sharp signal such as a Raman peak. Figure 2(b) shows the filter number dependence. As shown in the graph, the peak focusing became more accurate when the number of convolutional filters increased from 8 to 256. Otherwise, it seems the accuracy needed to find a feature for the test dataset was sufficient when 64 filters were used. Hence, the number of filters was set to 64 to reduce calculation time.

Next, we investigated the possibility of common component extraction with simple test spectra. Three random Lorentzian peaks were added to the training and test spectra in addition to the three defined peaks. One hundred spectra were prepared as the training dataset for each label (i.e. a total of 300 spectra). The example of the training spectra and additional explanation are shown in Fig. S2 (supporting information). Figure 3(a–c) shows the input test spectrum and its extracted feature. The highlight of the features extracted after the pooling (green solid lines) were calculated using the visual explanation approach expressed by Eqs. (6) and (7) (called the gradient-weighted class activation mapping [20]). On this layer, all Lorentzian peaks were extracted except for the defined common peak. The blue solid line shows the derived feature contributing to the recognition in our method. The extracted regions correspond well to the defined common peaks (100, 500, and 1,000 ch).

Consequently, we selected a relatively large number of filters with sizes close to the linewidth to extract Raman peaks for recognition and common component extraction using the proposed method.

*4.2. Feature extraction for pharmaceutical compound spectra*

Next, we applied the proposed method to measured Raman spectra from pharmaceutical compounds. Figure 4(a) shows representative spectra for employed pharmaceutical compounds. These examples have the highest signal-to-noise (SN) ratio in the obtained spectra. Forty-four spectra were obtained from each pharmaceutical compound, but the lens focus and measuring point at the sample were adjusted so that each spectrum had a different SN ratio. Four spectra with different SN ratios were selected as the training dataset and the others were used for the test dataset.

Examples of the spectra selected for the training dataset are shown in Fig. 4(b). We chose the spectrum with the highest signal, one with a signal that was approximately 1/2 of the highest one, and two with signals that were 1/3 and 1/4 as high. Wavenumber interpolation and normalization were applied to the spectra, as described in Section 2.1., before the datasets were input to the model.

The network was trained using stochastic gradient descent on the Adam optimizer with a learning rate of $1 \times 10^{-4}$ for 200 epochs using 5-fold cross validation.

Figure 5 shows the evaluated test spectra and their extracted features that contribute to the recognition. Note that the extracted features consist of the strong Raman peaks represented by the evaluated compounds, as shown in Fig. 5. The spectra for the LaE sample had especially low SN ratios, which were also reflected in the extracted features. Near-zero values and baseline flattening were obtained at background level regions, for example, and the effect of baseline correction appeared on almost all of the extracted spectra, especially in the Knh and Njt spectra, which have relatively weak Raman signals. Spectra expressed as extracted features should indicate weights for classifying a test dataset. Namely, if a spectrum has a strong peak near a peak of one on the other spectra, the values of its weights are reduced. In contrast, if a spectrum has a peak at a region that does not have peaks in other spectra, the values of the weights are increased in this region. Therefore, the spectral shapes of the extracted features in Fig. 5 roughly correspond to the spectra in Fig. 4(a), although the relative intensity of peaks are different. We also derived the actual data in the networks and showed the example in Fig. S3 (supporting information). On this dataset, a 100% classification accuracy was obtained. This is a slightly better accuracy than that of the result derived by a support vector machine (SVM) [26] method combined with PCA [27] using scikit-learn in Python shown as Fig. S4 (supporting information). In addition, we also derived a visualization result using Partial Least Squares (PLS) projection and variable importance on projection (VIP) as shown in Fig. S5 (supporting information) and confirmed the possibility of better extraction result using our method.

In the case of the discrimination of pharmaceutical compounds using Raman spectra, we could visually recognize each compound by its characteristic peaks in the spectra. For instance, the characteristic spectral shapes around 1,600 $cm^{-1}$ for Buf and at 360 $cm^{-1}$ for Knh may intuitively be chosen to identify them from other pharmaceutical compounds in this dataset. Similarly, the spectral shapes of the extracted features for Buf and Knh are strongly highlighted in these peak regions. The details of Knh spectrum around 360 $cm^{-1}$ is shown in Fig. S6 (supporting information). This result is an important point of our study, because our method demonstrates that Raman spectra coupled with a CNN enables the 'chemically guaranteed' discrimination of compounds. We also obtained good recognition and visualization results by using 3-convolutional layers model. However, we need additional discussion to show the details, because some label had a highlight on only background region when we used a deeper model, while 100% classification accuracy was obtained. We expected the CNN models could understand that the background region, which did not have strong peaks, was also an important feature for its recognition in this case.

### 4.3. Common component extraction

Finally, we demonstrate the possibility of common component extraction by evaluating numerically mixed spectra of amino acids. In this section, all training were generated with a numerical mixing calculation by picking two spectra from 20 pure spectra datasets, as shown in Fig. 6. A spectrum of one amino acid was mixed with that of another using a randomly selected ratio in the range 0.1–0.5, five times for each amino acid. Namely, we prepared 95 (19 × 5) training spectra for each amino acid, obtaining a final total of 1,900 (95 × 20) for the training spectra. Note that the training labels were encoded by one-hot vectors: the mix ratio was not given in the true label. All test spectra and labels, which also obtaining a total of 1900, were prepared with same calculation. Wavenumber interpolation and normalization were applied to the spectra, as described in Section 2.1. The network was trained using stochastic gradient descent on the Adam optimizer with a learning rate of $1 \times 10^{-3}$ for 30 epochs using 5-fold cross validation.

In this paper, we explain the results of the Ala-mixed spectra. Figure 7(a) shows the pure Ala spectrum. Figure 7(b–d) shows the extracted results for Ala-mixed Arg, His, and Leu spectra (for the other results, see Fig. S7 in supporting information). The mix ratio of input spectra was 1:0.45. In Fig. 7(b), although the input spectrum exhibits a strong peak at 850 $cm^{-1}$ attributable to Ala, the extracted component shows peaks only for Arg. These results indicate that the common component in the training dataset was suitably extracted by our method. Near-zero values were obtained at the background noise and weak peak region, as in the results of the pharmaceutical compound analysis. In the extraction results for the His and Leu spectra displayed in Fig. 7(c), (d), a negative peak was shown at the Ala peak around 850 $cm^{-1}$. Hence, it is expected that the learning in our method recognizes the peak at 850 $cm^{-1}$ as important for the recognition of His and Leu in the Ala-mixed spectrum. The relatively weak peaks of pure Leu at 470 $cm^{-1}$ and 1,470 $cm^{-1}$ were strongly highlighted in the extracted feature spectra because the peak at 850 $cm^{-1}$ was very close to the strongest peak of Leu (at 850 and 861 $cm^{-1}$ after interpolation). The reasons for this phenomenon were discussed in the previous section. On this dataset, we also obtained 100% classification accuracy, which is slightly better than that of the SVM combined with PCA method (see Fig. S8 in supporting information). Common component and extraordinary peak extractions using our proposed method were also demonstrated by employing one-hot vectors for the labels of the mixed spectra.

## 5. Conclusions

In this paper, we demonstrated a recognition and feature visualization method based on CNN for evaluating Raman spectra. The proposed method derives "importance weights" from the FC layer using the gradients of a target label. For the extraction of a Raman-peak signal, we determined the optimal filter size (close to the linewidth) and number of filters using numerically generated Lorentzian spectra. The extraction was also confirmed by the evaluation of experimentally observed Raman spectra. The extracted spectra had strong highlights at strong Raman-peak regions. In addition, near-zero values and baseline flattening were obtained at the background level region, which appeared to be caused by the baseline flattening effect.

Common component extraction from numerically mixed spectra was also evaluated. Although a one-hot vector (without the mix ratio) was given as the true training label, recognition and common component extraction were demonstrated. This proposed method should be usable for the validation of a trained model, ensuring the reliability of common component extraction from compound samples for spectral analysis.

**Appendix A. Supplementary data**

Supplementary data to this article can be found online at   https://doi.org/10.1016/j.aca.2019.08.064.

**REFERENCES**

[1] H. Mansour and A. Hickey, Raman Characterization and Chemical Imaging of Biocolloidal Self-Assemblies, Drug Delivery Systems, and Pulmonary Inhalation Aerosols; A Review, *AAPS Pharm. Sci. Tech.* **8**, (2007) 140-E16.

[2] E. Hanlon, R. Manoharan, T. Koo, K. Shafer, J. Motz, M. Fitzmaurice, J. Kramer, I. Itzkan, R. Dasari and M. Feld, Prospects for in vivo Raman spectroscopy, *Phys. Med. Biol.* **45**, (2000) R1-R59.

[3] J. A. Dieringer, A. D. McFarland, N. C. Shah, D. A. Stuart, A. V. Whitney, C. R. Yonzon, M. A. Young, X. Zhang and R. P. V. Duyne, Surface enhanced Raman spectroscopy: new materials, concepts, characterization tools, and applications, *Faraday Discuss.* **132,** (2006) 9-26.

[4] T. Bocklitz, A. Walter, K. Hartmann, P. Rosch and J. Popp, How to pre-process Raman spectra for reliable and stable models? *Anal. Chim. Acta.* **704**, (2011) 47-56.

[5] N. K. Afseth, V. H. Segtnan and J. P. Wold, Raman Spectra of Biological Samples: A Study of Preprocessing Methods, *Appl. Spectrosc.* **60**, (2006) 1358-1367.

[6] G. Schulze, A. Jirasek, M. Yu, A. Lim, R. Turner and M. Blades, Investigation of Selected Baseline Removal Techniques as Candidates for Automated Implementation, *Appl. Spectrosc.* **59**, (2005), 545-574.

[7] P. J. Grood, G. J. Postma, W. J.Melssen, L. M. C. Buydens, V. Deckert and R. Zenobi, Application of principal component analysis to detect outliers and spectral deviations in near-field surface-enhanced Raman spectra, *Anal. Chim. Acta* **446**, (2001) 71-83.

[8] S. Sigurdsson, P. Philipsen, L. Hansen, J. Larsen, M. Gniadecka and H. Wulf, Detection of Skin Cancer by Classification of Raman Spectra, *IEEE Trans. Biomed. Eng.* **10**, (2004) 1784-1793.

[9] S. Feng, D. Lin, J. Lin, B. Li, Z. Huang, G. Chen, W. Zhang, L. Wang, J. Pan, R. Chen and H. Zeng, Blood plasma surface-enhanced Raman spectroscopy for non-invasive optical detection of cervical cancer, *Analyst* **138** (2013), 3967-3974.

[10] Y. LeCun, Y. Bengio and G. Hinton, Deep learning, *Nature* **521**, (2014) 436-444.

[11] A. Krizhevsky, I. Sutskever and G. E. Hinton, ImageNet Classification with Deep Convolutional Neural Networks, In *Proc. Advances in Neural Information Processing Systems* **25**, (2012) 1090-1098.

[12] O. Russakovsky, J. Deng, H. Su, J. Krause, S. Satheesh, S. Ma, Z. Huang, A. Karpathy, A. Khosla, M. Bernstein, A. C. Berg, L. Fei, ImageNet Large Scale Visual Recognition Challenge *Int. J. Comput. Vis.* **115**, (2015) 211-252.

[13] C. Dong, C. C. Loy, K. He and X. Tang, Image Super-Resolution Using Deep Convolutional Networks, *IEEE Trans. Patt. Anal. Mach. Int.* **38**, (2016) 295-307.

[14] T. N. Sainath, A. Mohamed, B. Kingsbury and B. Ramabhadran, Deep convolutional neural networks for LVCSR, *Proc. Acoustics, Speech and Signal Processing*, 2013, 8614–8618.

[15] C. Ni, D.Wang and Y. Tao, Variable weighted convolutional neural network for the nitrogen content quantization of Masson pine seedling leaves with near-infrared spectroscopy, *Spectrochim. Acta. Pt. A. Mol. Bio.* **209**, (2019) 32-39.

[16] J. Acquarelli, T. Laarhoven, J. Gerretzen, T. N. Tran, L. M. C. Buydens and E. Marchiori, Convolutional neural networks for vibrational spectroscopic data analysis, *Anal. Chim. Acta* **954**, (2017) 22-31.

[17] J. Liu, M. Osadchy, L.Ashton, M. Foster, C. J. Solomon and S. J. Gibson, Deep Convolutional Neural Networks for Raman Spectrum Recognition: A Unified Solution, *Analyst* **21**, (2018) 4067-4074.

[18] W. Zhang and S. Zhu, Visual interpretability for Deep Learning: a Survey, *Front. Inf. Technol. Electron. Eng.* **19**, (2018) 27–39.

[19] D. Smilkov, N. Thorat, B. Kim, F. Viegas and M. Wattenberg, "Smooth Grad: removing noise by adding noise," *arXiv*: 1706.03825v1, 2017

[20] R. R. Selvaraju, M. Cogswell, A. Das, R. Vedantam, D. Parikh, D. Batra, Grad-CAM: Visual Explanations from Deep Networks via Gradient-based Localization, *arXiv*: 1610.02391v3, 2017.

[21] M. Abadi, A. Agarwal, P. Barham, E. Brevdo, Z. Chen, C. Citro, G. S. Corrado, A. Davis, J. Dean, M. Devin, S. Ghemawat, I. Goodfellow, A. Harp, G. Irving, M. Isard, Y. Jia, R. Jozefowicz, L. Kaiser, M. Kudlur, J. Levenberg, D. Mane, R. Monga, S. Moore, D. Murray, C. Olah, M. Schuster, J. Shlens, B. Steiner, I. Sutskever, K. Talwar, P. Tucker, V. Vanhoucke, V. Vasudevan, F. Viegas, O. Vinyals, P. Warden, M. Wattenberg, M. Wicke, Y. Yu and X. Zheng, TensorFlow: Large-Scale Machine Learning on Heterogeneous Distributed Systems, *arXiv*: 1603.04467, 2016.

[22] A. L. Maas, A. Y. Hannun, and A. Y. Ng, Rectifier nonlinearities improve neural network acoustic models, *Proc. ICML*, 2013.

[23] N. Srivastava, G. hinton, A. Krizhevsky, I. Sutskever and R. Salakhutdinov, Dropout: A Simple Way to Prevent Neural Networks from Overfitting, *J. Mach. Learn. Res.*, **15**, (2014) 1929-1958.

[24] J. L. Elman, Finding Structure in Time, *Cogn. Sci. **14***, (1990) 179-211.

[25] K. Simonyan and A. Zisserman, Very Deep Convolutional Networks for Large-Scale Image Recognition, *Proc. ICLR*, (2014).

[26] C. Cortes and V. Vapnik, Support-vector Networks, *Mach. Learn.* **20**, (1995), 273–297.

[27] S. Wold, K. Esbensen, and P. Geladi, Principal component analysis, *Chemom. Int. Lab. Syst.* **2**, (1987) 37-52.

Analytica Chimica Acta 00 (2019) 000–000  9

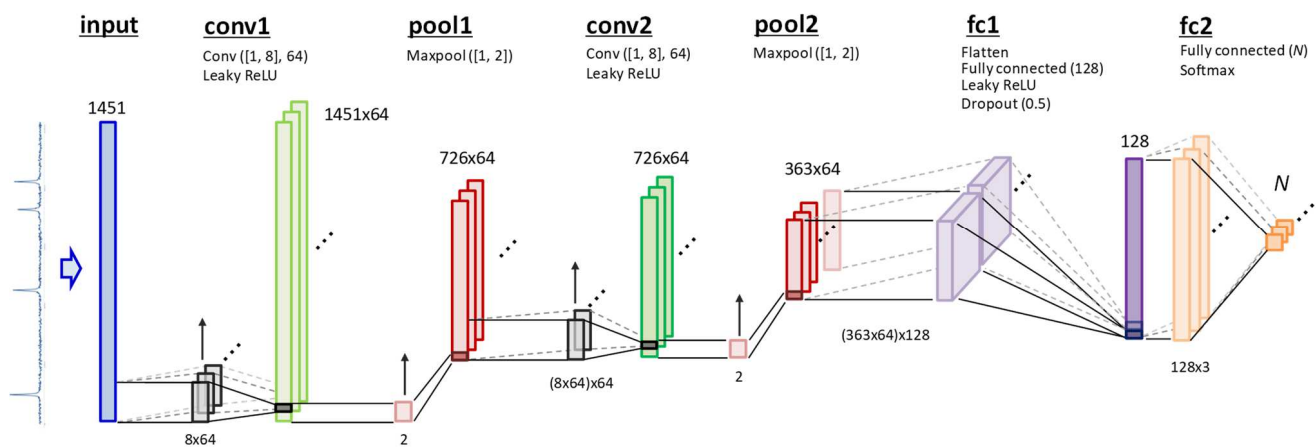

Figure 1: CNN model for Raman spectrum analysis, where *N* is the number of classes. Here, Conv ([1, 8], 64) denotes a convolutional layer with 64 filters of size 1 × 8 (with a stride size of 1). Maxpool ([1,2]) denotes a max pooling layer with one filter of size 1 × 2. Flatten denotes a layer that convert their input to 1D data for input of following fully connected layer. Fully connected (128) denotes a fully-connected layer with 128 outputs. Dropout (0.5) denotes a layer that randomly dropout 50% neurons of the previous layer during training.

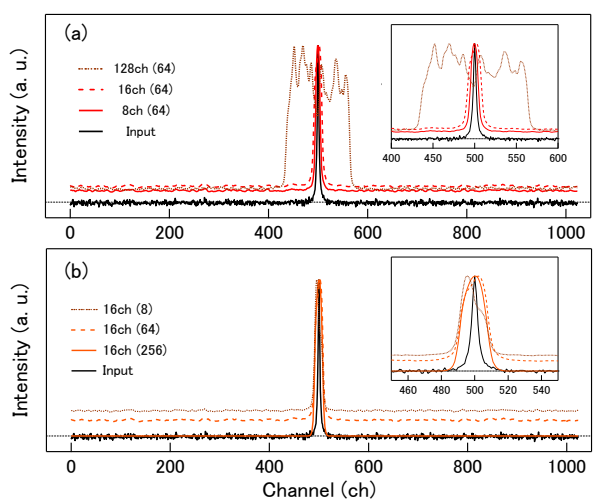

Figure 2: Examples of generated test spectra (input) and extracted features. Applied filter size (ch) and the number of filter (in brackets) are described in the legends. (a) Filter size dependence of the extracted feature. The inset shows an enlarged graph in the range 400–600 ch. (b) Filter number dependence of the extracted feature. The inset shows an enlarged graph in the range 450–550 ch.

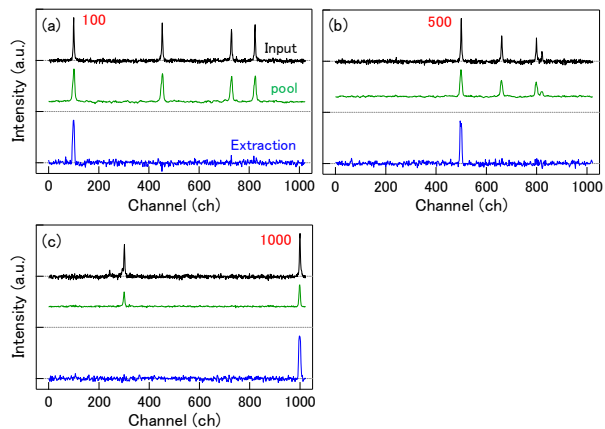

Figure 3: Common component extraction with the spectrum including additional random Lorentzian peaks. Trained common peak positions are defined at (a) 100 ch, (b) 500 ch, and (c) 1,000 ch. Black, green, and blue solid lines indicate the input test spectrum, feature extracted after the pooling layer, and extracted feature contributing to the recognition with proposed method, respectively.

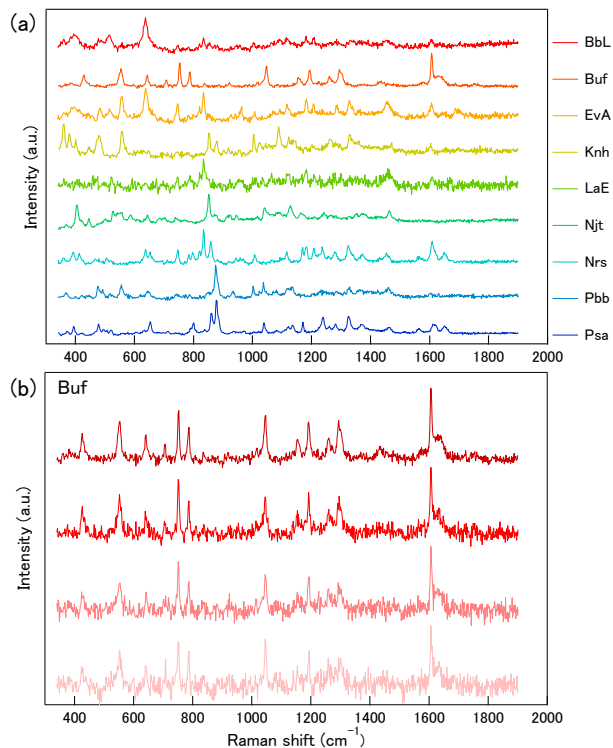

Figure 4: (a) Representative spectra for the evaluated pharmaceutical compounds. (b) Examples of selected spectra used in the training datasets.

Analytica Chimica Acta  00 (2019) 000–000       
**11**

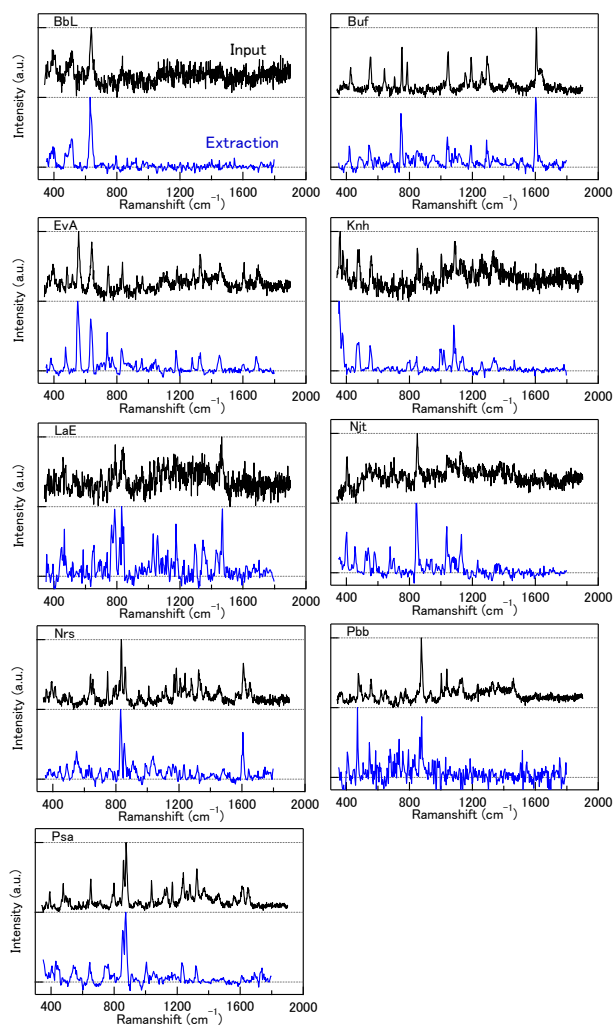

Figure 5: Features extracted from the pharmaceutical compound spectra. The black and blue solid lines show the input test spectra and extracted feature contributing to the recognition, respectively. The intensity was normalized relative to each maximum value.

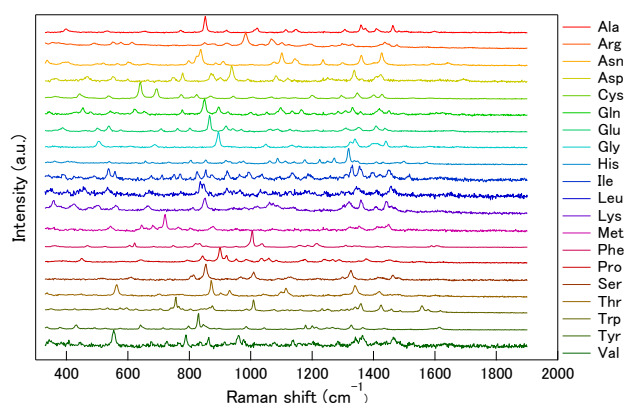

Figure 6: Prepared pure spectra measured from amino acids for the evaluation of common component extraction.

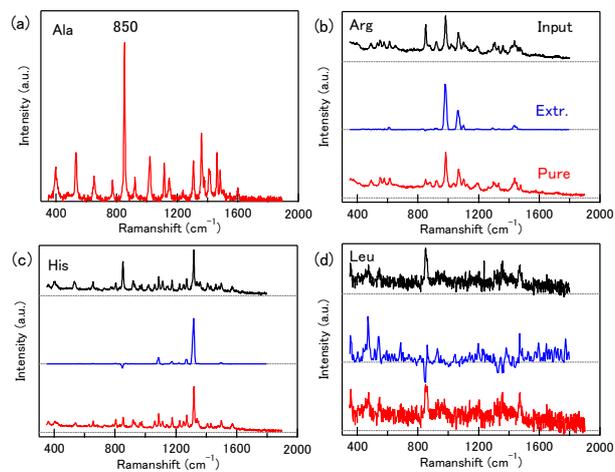

Figure 7: Examples of common component extraction from the compound spectrum. (a) Spectrum measured from the pure Ala sample. (b)–(d) Evaluation results of Ala-mixed Arg, His, and Leu spectra. Inputs, extracted features, and each pure spectrum are indicated by black, blue, and red solid lines, respectively.